\documentclass{article} 
\usepackage{iclr2021_conference,times}

\usepackage{amsmath,amsfonts,bm}

\def\eqref#1{equation~\ref{#1}}

\def\1{\bm{1}}

\DeclareMathAlphabet{\mathsfit}{\encodingdefault}{\sfdefault}{m}{sl}
\SetMathAlphabet{\mathsfit}{bold}{\encodingdefault}{\sfdefault}{bx}{n}

\usepackage{hyperref}
\usepackage{url}
\usepackage{graphicx}
\usepackage{subfig} 

\title{PyVertical: A Vertical Federated Learning Framework for Multi-headed SplitNN}

\author{%
    Daniele~Romanini \thanks{ Authors have equal contribution in this work}\\
    OpenMined\\
    \texttt{daler.romanini@gmail.com}
    \And
    Adam~James~Hall $^{*}$\\
    Edinburgh Napier University / OpenMined\\
    \texttt{adam@openmined.org} 
    \AND
    Pavlos~Papadopoulos $^{*}$\\
    Edinburgh Napier University / Apheris\\
    \texttt{pavlos.papadopoulos@napier.ac.uk} 
    \And
    Tom~Titcombe $^{*}$ \\
    Tessella / OpenMined\\
    \texttt{tom.titcombe@tessella.com~~~} 
    \AND
    Abbas~Ismail\\
    Birla Institute of Technology, Mesra\\
    \texttt{be10285.17@bitmesra.ac.in} 
    \And
    Tudor~Cebere\\
    OpenMined\\
    \texttt{tudor@openmined.org~~~~~~~~~} 
    \AND
    Robert Sandmann\\
    Apheris\\
    \texttt{r.sandmann@apheris.com} 
    \And
    Robin~Roehm\\
    Apheris\\
    \texttt{r.roehm@apheris.com~~~~~~~~~} 
    \AND
    Michael~A.~Hoeh\\
    Apheris \\
    \texttt{m.hoeh@apheris.com} 
}

\iclrfinalcopy 
\begin{document}

\maketitle

\begin{abstract}
We introduce PyVertical, a framework supporting vertical federated learning using split neural networks. The proposed framework allows a data scientist to train neural networks on data features vertically partitioned across multiple owners while keeping raw data on an owner's device. To link entities shared across different datasets' partitions, we use Private Set Intersection on IDs associated with data points. To demonstrate the validity of the proposed framework, we present the training of a simple dual-headed split neural network for a MNIST classification task, with data samples vertically distributed across two data owners and a data scientist.


\end{abstract}

\section{Introduction}

With ubiquitous data collection, individuals are constantly generating diverse swathes of data, including location, health, financial information. These data streams are often collected by separate entities and are sufficient for high utility use-cases. A common challenge faced by data scientists is utilising data isolated in silos to train machine learning (ML) algorithms. When this data is commercially sensitive, personal or otherwise under strict legal protection, it cannot be simply merged with data controlled by another party. To ensure data privacy is not compromised during the training or inference process, several privacy-preserving ML techniques, such as Federated Learning (FL) \citep{mcmahan2016communication,konevcny2016federated,mcmahan2017federated,bonawitz2019towards,ryffel2018generic}, focus on training ML models on distributed datasets by keeping data in the custody of its corresponding holder. FL typically splits data horizontally. This is where datasets are distributed across multiple owners that have the same features and represent different data subjects \citep{kantarcioglu2004privacy}. However, it is common in real-world scenarios to find datasets which are vertically distributed \citep{mcconnell2004building}, i.e. different features of the same data subject are distributed across multiple data owners. For example, specialists or general hospitals may hold different parts of a patient's medical data.

To address the issue of learning from vertically distributed data, we use Split Neural Networks (SplitNN) to first map data into an abstract, shareable representation. This allows information from multiple sources to be combined for learning without exposing raw data. We combine this with Private Set Intersection (PSI) to identify and link data points belonging to the same data samples shared among parties. This process facilitates Vertical Federated Learning (VFL) for non-linear functions.

\subsection{Contributions}
\label{contributions}

In this work, we extend the proposal of \citep{angelou2020asymmetric}, regarding the use of (SplitNNs) and PSI in Vertical Federated Learning. We use the PySyft library for privacy-preserving machine learning \citep{ryffel2018generic} to train a Vertically Federated ML algorithm on data distributed across the premises of one or multiple data owners. This work is released as an open-source framework, PyVertical. To the best of our knowledge, this is the first open-source framework to perform machine learning on vertically distributed datasets using Split Neural Networks\footnote{Code is available at PyVertical:\href{https://github.com/OpenMined/PyVertical}{ https://github.com/OpenMined/PyVertical}}. 

We verify our method on a two-party, vertically-partitioned MNIST dataset. Our work presents a dual-headed scenario, where data from two separate data owners (who holds different parts of the data samples) and a data scientist (who, in our case, holds data labels) are securely aligned and combined for model training. However, this work could be extended to multiple data owners using the same principle we describe here. 


\section{Background Knowledge and Related Work} 
\label{litreview}

\subsection{Private Set Intersection}
\label{PSI}


Private Set Intersection (PSI) \citep{freedman2004efficient,huang2012private,de2010practical,dachman2009efficient} is a multi-party computation cryptographic technique which allows two parties, where each hold a set of elements, to compute the intersection of these elements, without revealing anything to the other party except for the elements in the intersection. Different PSI protocols have been proposed \citep{buddhavarapu2020private,ion2020deploying, chase2020private,pinkas2018scalable} and employed for scenarios such as private contact discovery~\citep{demmler2018contactdiscovery} and also privacy-preserving contact tracing \citep{angelou2020asymmetric}. 

In this work, we use a PSI implementation based on a Diffie-Hellman key exchange that uses Bloom filters compression to reduce the communication complexity~\citep{angelou2020asymmetric}. This protocol works with two parties computing the intersection between their sets. However, the chosen PSI framework can be replaced with an alternative implementation, for instance, to compute directly the intersection of datasets coming from more than two parties~\citep{hazay2017scalable}.

\subsection{Split Neural Networks}
\label{splitNNreview}

Split learning is a concept of training a model that is split into segments held by different parties or on different devices. A neural network model trained this way is called a Split Neural Network, or SplitNN. In SplitNN, each model segment transforms its input data into an intermediate data representation (as the output of a hidden layer of a classic neural network). This intermediate data is transmitted to the next segment until the training or the inference process is completed. During backpropagation, the gradient is also propagated across different segments. Compared to data-centric FL, split learning can also be useful to reduce the computational burden on data owners, who in many real-world scenarios may have limited computational resources~\citep{gupta2018distributed,vepakomma2018split}.

\subsection{Vertical Federated Learning}
\label{vfl}

Vertical federated learning (VFL) is the concept of collaboratively training a model on a dataset where data features are split amongst multiple parties \citep{yang2019federated}. For example, different healthcare organizations may have different data for the same patient. Considering the sensitivity of the data, these two organizations cannot simply merge their information without violating that person's privacy. For this reason, a machine learning model should be trained collaboratively, and data should be kept on the corresponding premises.
Machine learning algorithms for vertical partitioned data is not a new concept, and many studies for new models and algorithms have been proposed in this area \citep{feng2020multi,liu2020asymmetrically,du2001privacy,du2004privacy,vaidya2002privacy,karr2009privacy,sanil2004privacy,wan2007privacy,gascon2017privacy,thapa2020splitfed, hardy2017private, nock2018entity}. Existing open-source VFL frameworks include FedML \citep{he2020fedml}, which implements multi-party linear models~\citep{hardy2017private}.

Similarly to our work, the use of split networks for vertical federated learning has been proposed~\citep{ceballos2020splitnndriven}. However, differently from our work, the authors investigate multiple methods for combining information sent to the data scientist from disjoint datasets. Moreover, they do not consider the entity resolution problem for aligning data across parties, whereas we illustrate how PSI can be successfully exploited prior to the training process to account for this.

\section{Framework Description}
\label{frameworkdescription}

We introduce PyVertical, a framework written in Python for vertical federated learning using SplitNNs and PSI. PyVertical is built upon the privacy-preserving deep learning library PySyft~\citep{ryffel2018generic} to provide security features and mechanisms for model training, such as pointers to data, without exposing private information.

A set of data features are distributed across one or more data owners. We refer to a full dataset split vertically across the features as a \textit{vertical dataset}. Each of the data owners takes part in the model training, alongside a data scientist who orchestrates the process. The data scientist could also be a data owner itself, holding features or data labels. The data features in the vertical datasets may intersect. Each data point is associated with a unique ID based on the data point's subject, the format of which is agreed by the data owners (e.g. legal names, email addresses, ID card numbers). The data owners use PSI to agree on a shared set of data IDs (process described in Section \ref{psi-protocol}); each data owner discards non-shared data from their datasets and sorts their datasets by ID, such that element $n$ of each vertical dataset corresponds to the same data subject.

\begin{figure}[!h]
  \centering
  \includegraphics[width=0.65\linewidth]{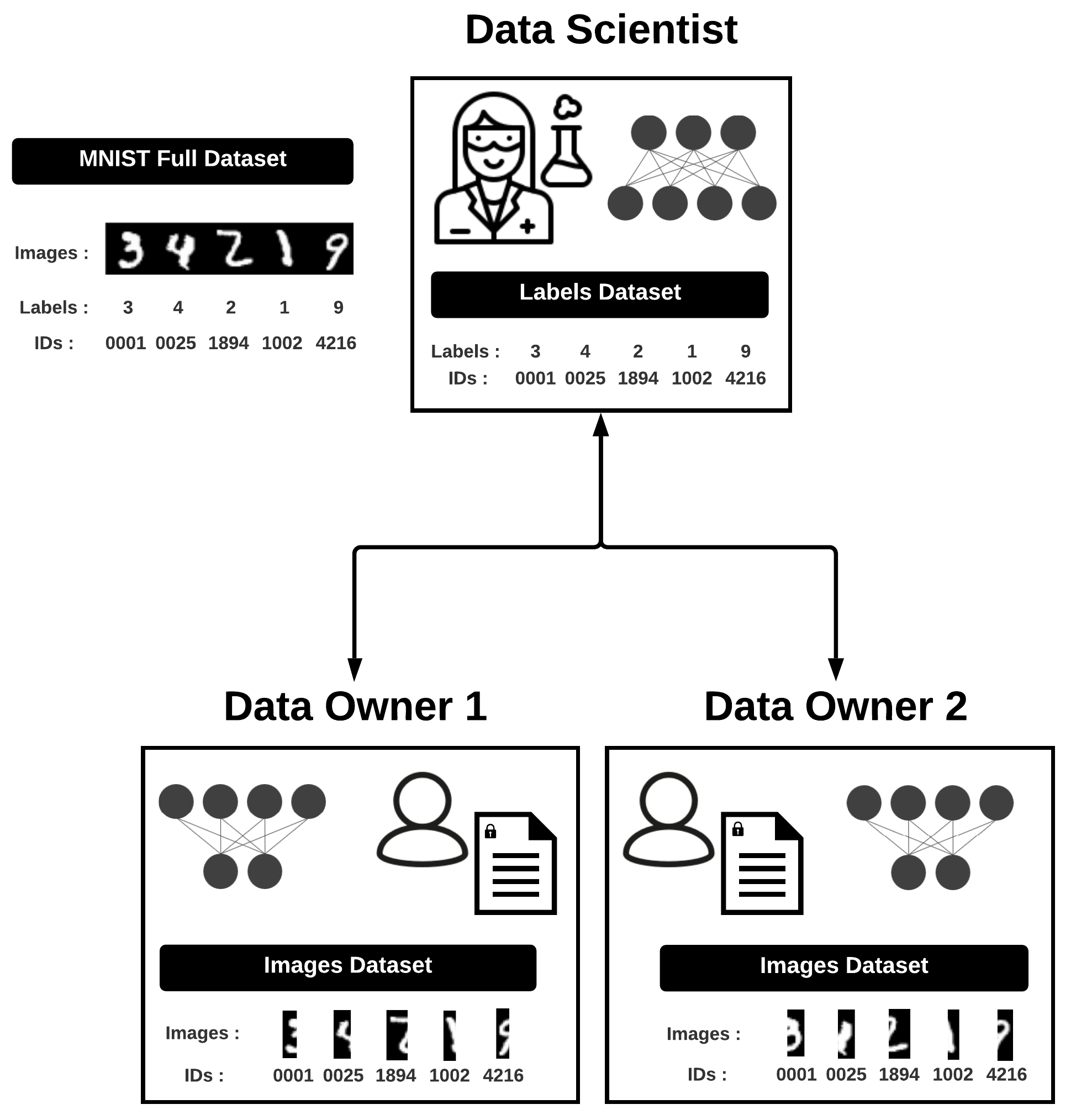}
  \caption{Parties and datasets in the conducted experiment. Data Scientist holds a part of the SplitNN and the labels dataset. Data Owners hold their images datasets and parts of the SplitNN}
  \label{fig:pyvertical-architecture}
\end{figure}

In our framework, the data scientist is able to define a split neural network model and send the corresponding model segments to the data owners.
Each data owner's model segment maps their data samples to an abstract representation with $k_i$ neurons. The output from each model segment (which would correspond to a hidden layer of a complete classic neural network) is sent to the data scientist and concatenated to form a $\sum_i k_i$ length vector. The data scientist also defines a model segment corresponding to the final part of the split neural network. This segment remains on the data scientist's premises and maps the concatenated, intermediate data (i.e. the output from data owners' model segments) into a shape relevant to the task. During model training, the data scientist calculates batch loss and updates their model segment's weights. The data scientist then sends the final gradient to the data owners, each of whom updates their own model segment by completing backpropagation independently. We assume all parties are honest-but-curious actors. Figure~\ref{fig:pyvertical-architecture} demonstrates model inference under this framework for the experiment outlined in Section~\ref{experiment}.

\subsection{Data Resolution Protocol}
\label{psi-protocol}

We use a PSI Python library \citep{PSIsourcecode} to identify intersections between data points in two datasets based on unique IDs. In this work, we consider a setting where the data scientist has access to ground truth labels. For all three parties (two data owners + one data scientist) to agree on data points shared among all datasets, the protocol works as follow: firstly, the data scientist runs the PSI protocol independently with each data owner. The intersection of IDs between the data scientist and each data owner is revealed to the data scientist. The data scientist computes the global intersection from the two previous independently computed intersections and communicates the global intersection to the data owners. In this setting, the data owners do not communicate and are not aware of each other's identity in any regard. In practice, this is an ideal feature of the protocol as having knowledge of a group's or individual's participation in a training process can reveal sensitive commercial and personal information in and of itself. Moreover, as the single IDs' intersection lists are only revealed to the data scientist, there is no risk for the data owners to learn which information the other data owners owns. Each of the data owners learns only the information necessary for training or inference. 

\section{Experiment}
\label{experiment}

To verify the validity of our framework, we train a dual-headed SplitNN on a vertically-partitioned version of the MNIST dataset. We generate the data by splitting the images in MNIST into left and right halves, providing a dataset of each half to different data owners. The data scientist defines and sends an identical, multi-layered neural network to each of the data owners that takes 392-length vectors as input (flattened representation of 28x14 pixel images). The data scientist also defines and keeps on its premises the second part of the neural network, which outputs a softmax layer for classification. The data scientist can access the ground truth labels and calculate the loss for each data batch. The data scientist controls the training process and hyperparameters. Appendix~\ref{app:experiment} provides more details on the specific values used in model training.
The objective of this experiment is to demonstrate that the proposed framework allows vertically-partitioned learning. This specific experiment should be considered a proof-of-concept, thus not highly optimised for the specific task. Nevertheless, we report the results of the experiment in Figure~\ref{fig:train-results} (in Appendix~\ref{app:experiment}).

\section{Evaluation and Conclusion}
\label{evalconclusions} 

We have developed and distributed our work as an open-source project. We hope that PyVertical can serve as a useful tool for researching neural-networks-based VFL. We find PSI an appropriate and useful method for resolving data subjects across datasets; many datasets and domains already collect unique IDs, such as usernames or national identifiers for medical data, making our method widely applicable. Finally, we successfully train a dual-headed model on a vertically-partitioned MNIST dataset, demonstrating that the proposed framework and method work in principle.

\subsection{Limitations and Future Work}
\label{future-work}

The experiment performed in this work assumes that all the parties involved (data owners and data scientist) act honestly. To develop a truly scalable, robust VFL system, additional precautions should be taken into account: identity management, validation of adherence to PSI protocol, and a method agreeing on data ID schema, to name a few.

This work investigates a symmetric SplitNN model: we assume that each data owner holds an identical model segment and that data points are split equally between data owners. Future work should investigate the impact of imbalanced vertical datasets~\citep{liu2020asymmetrically} and the resulting difficulties from the asymmetric model segment convergence due to the use of different sized models and learning rates. 

Finally, we illustrate an example of a training process with two data owners and a data scientist holding labels. While the proposed framework can support more parties in principle, we aim to investigate how to apply the process to massively multi-headed Vertical Federated Learning tasks. Additionally, we plan to research and integrate other privacy-preserving ML techniques into our workflow, such as decentralised identities \citep{papadopoulos2021privacy,abramson2020distributed} and differential privacy \citep{dwork2006dp,dwork2008dp,titcombe2021practical}, to further enhance privacy guarantees.




\bibliography{iclr2021_conference}
\bibliographystyle{iclr2021_conference}

\appendix
\section{PyVertical Protocol}
\label{app:a}

\begin{figure}[h]
    \centering
    \subfloat[Full dataset]{{\includegraphics[width=5.6cm]{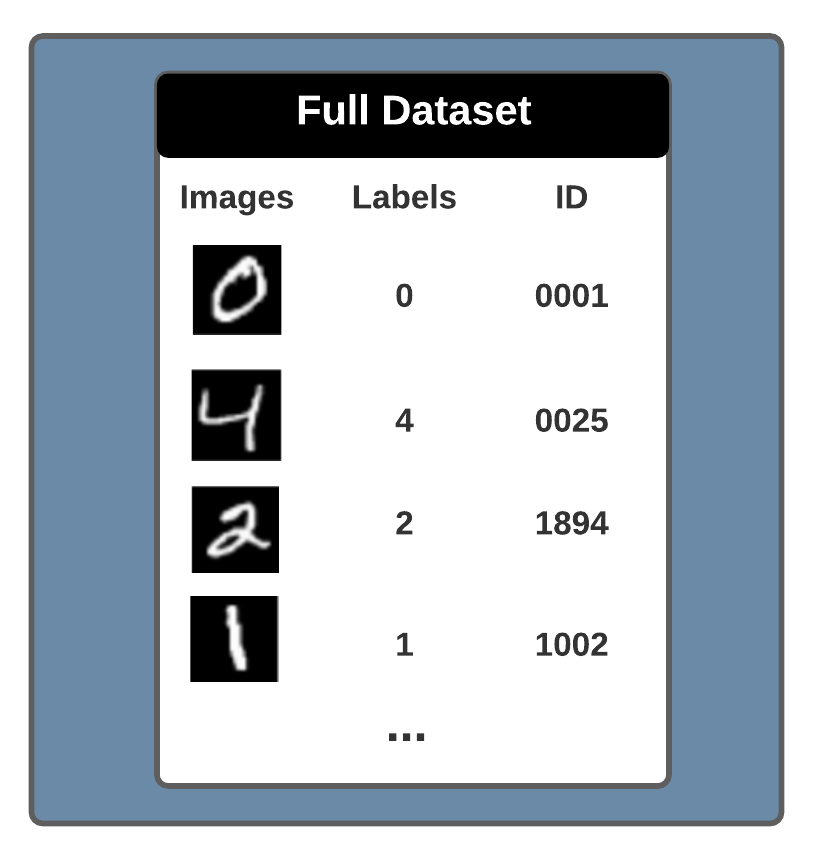} }}%
    \qquad
    \subfloat[Split images and labels datasets]{{\includegraphics[width=7.1cm]{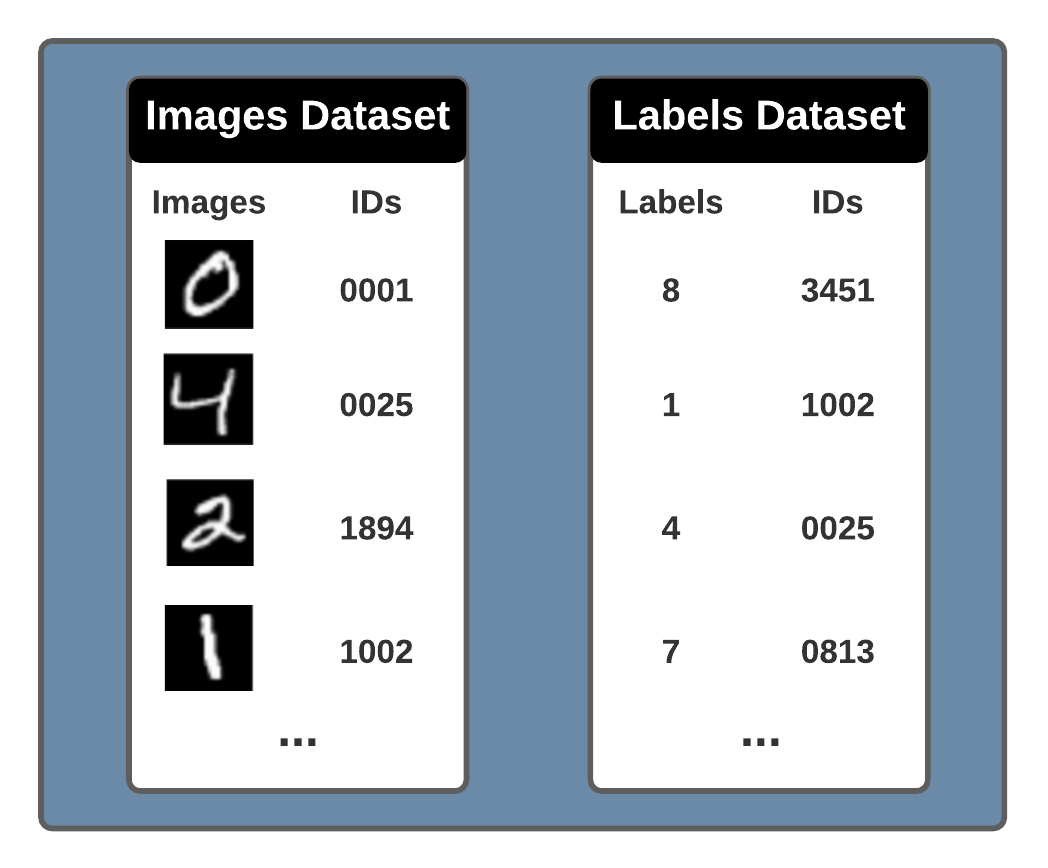} }}
    \qquad 
    \subfloat[PSI linkage and ordering]{{\includegraphics[width=6cm]{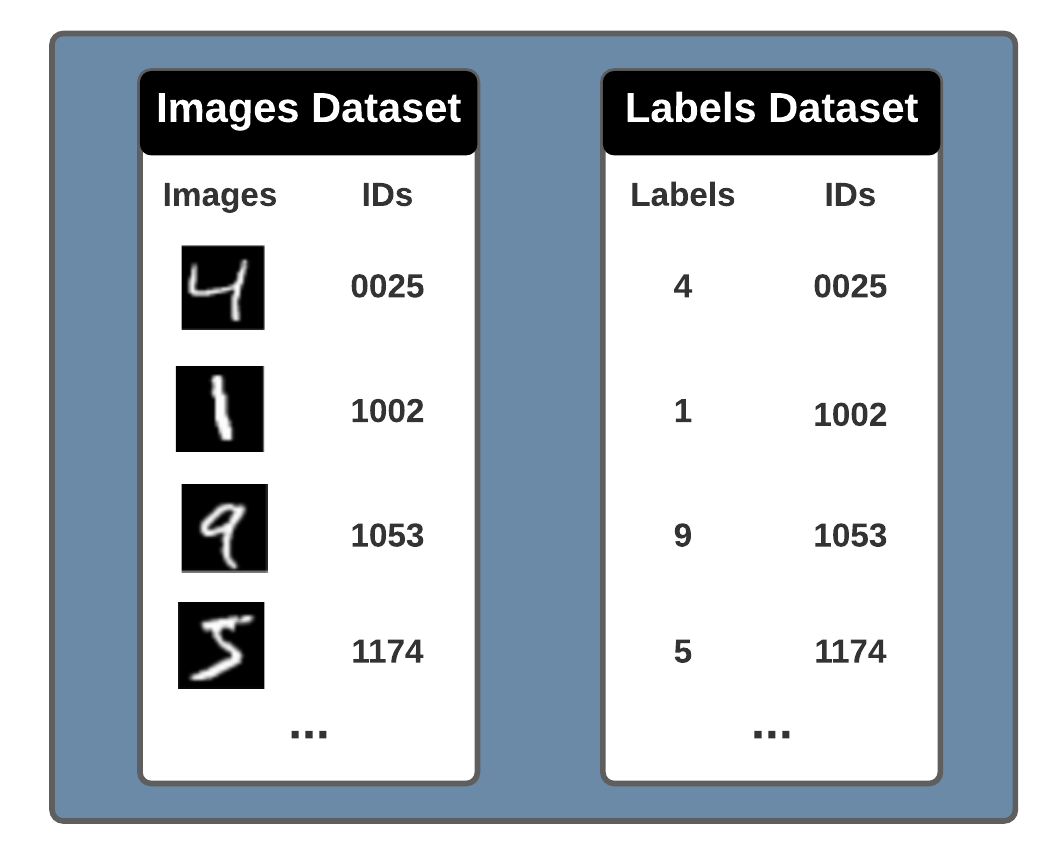} }}%
    \qquad
    \subfloat[SplitNN training]{{\includegraphics[width=7cm,height=4.85cm]{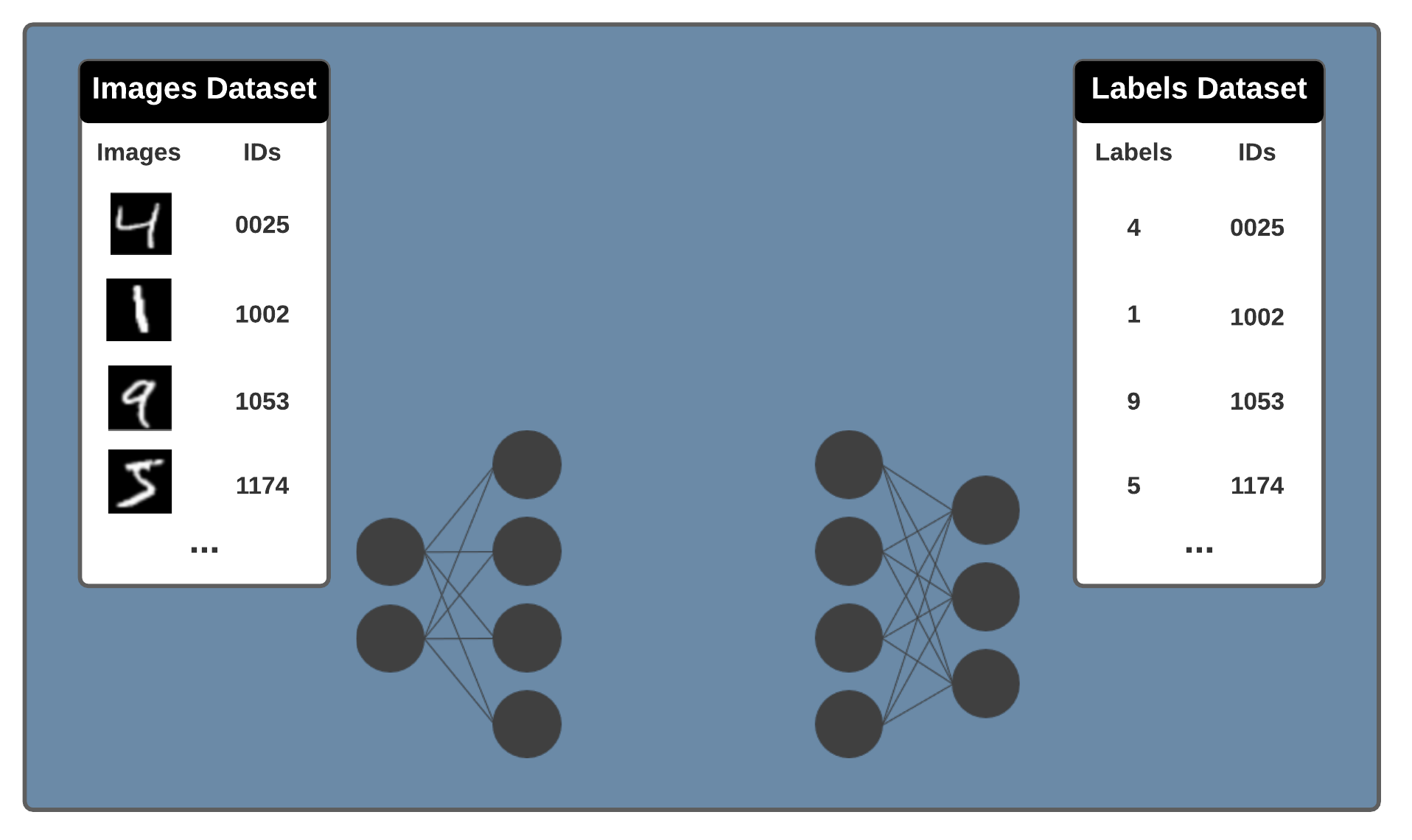} }}
    \caption{Vertical federated learning proof-of-concept implementation of \cite{angelou2020asymmetric}}
    \label{fig:pyvertical}
\end{figure}

\begin{figure}[!h]
  \centering
  \includegraphics[width=0.9\linewidth]{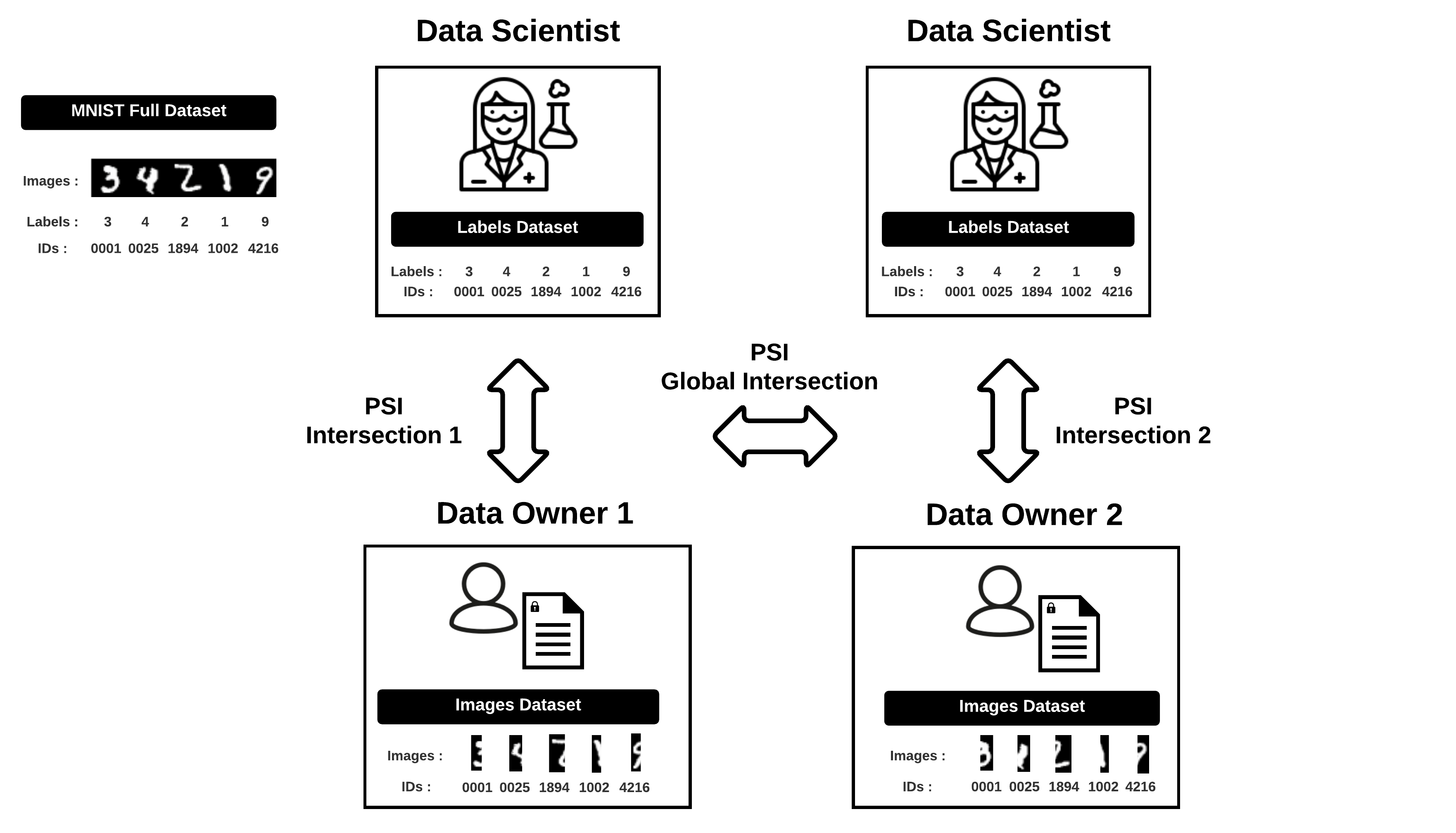}
  \caption{i) Data Scientist computes the intersection with Data Owner 1. ii) Data Scientist computes the intersection with Data Owner 2. iii) Data Scientist computes the global intersection.}
  \label{fig:pyvertical-architecture-expanded}
\end{figure}

Figure~\ref{fig:pyvertical} describes the PyVertical protocol applied to the MNIST dataset for a single data owner. The dual-headed PSI data linkage process is presented in Figure~\ref{fig:pyvertical-architecture-expanded}. Note that, in this illustration, there is only one data scientist; the duplicated icon is just to illustrate in more details how the data scientist runs a single PSI with each data owner separately, and that this could be done in parallel. 

\section{Experimental Setup}
\label{app:experiment}

The data owner model segment maps 392-length input into a 64-length intermediate vector with a ReLU activation, which is an abstract encoding of the data. The data scientist controls a separate neural network that takes as input a 128-length vector (concatenated data owner outputs) and transforms it into a softmax-activated, 10-class vector representing the possible digits in the dataset. The data scientist's model has a 500-length hidden layer with a ReLU activation. All layers are fully-connected. A learning rate of 0.01 is used for the data owner models and 0.1 for the data scientist model. Data is grouped into batches of size 128. Only the first 20,000 training images of MNIST are used, and the model is trained for 30 epochs.

\begin{figure}[htb]
  \centering
  \includegraphics[width=0.8\linewidth]{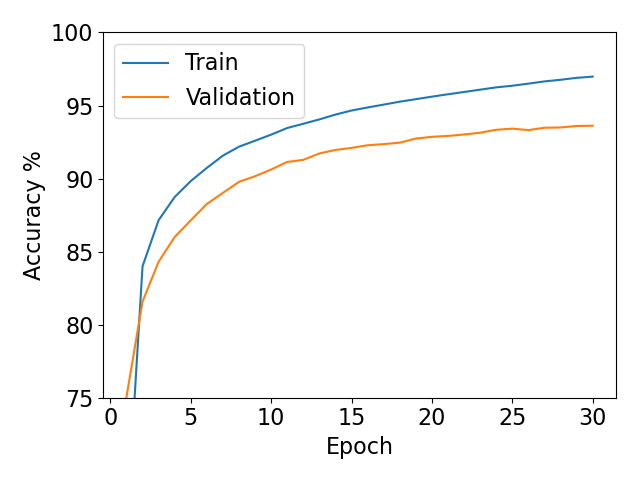}
  \caption{Train and validation accuracy for an unoptimised dual-headed SplitNN on vertically-partitioned MNIST.}
  \label{fig:train-results}
\end{figure}

\end{document}